%% file: sample-sigconf.tex
\documentclass[sigconf]{acmart}

\AtBeginDocument{%
  \providecommand\BibTeX{{%
    \normalfont B\kern-0.5em{\scshape i\kern-0.25em b}\kern-0.8em\TeX}}}

\setcopyright{acmcopyright}
\copyrightyear{2021}
\acmYear{2021}
\setcopyright{rightsretained}
\acmConference[MM '21]{Proceedings of the 29th ACM International
Conference on Multimedia}{October 20--24, 2021}{Virtual Event, China}
\acmBooktitle{Proceedings of the 29th ACM International Conference on
Multimedia (MM '21), October 20--24, 2021, Virtual Event,
China}
\acmISBN{978-1-4503-8651-7/21/10}
\acmDOI{10.1145/3474085.3478870}

\begin{document}
\fancyhead{} 

\title{Image Quality Assessment in the Modern Age}


\author{Kede Ma}
\affiliation{%
  \institution{City University of Hong Kong}
  \country{Hong Kong SAR, China}
  }
\email{kede.ma@cityu.edu.hk}


\author{Yuming Fang}
\affiliation{%
 \institution{Jiangxi University of Finance and Economics}
 \country{Jiangxi, China}
 }
\email{fa0001ng@e.ntu.edu.sg}

\renewcommand{\shortauthors}{Trovato and Tobin, et al.}

\begin{abstract}
  This tutorial provides the audience with the basic theories, methodologies, and current progresses of image quality assessment (IQA). From an actionable perspective, we will first revisit several subjective quality assessment methodologies, with emphasis on how to properly select visual stimuli. We will then present in detail the design principles of objective quality assessment models, supplemented by  an in-depth analysis of their advantages and disadvantages. Both hand-engineered and (deep) learning-based methods will be covered. Moreover, the limitations with the conventional model comparison methodology for objective quality models will be pointed out, and novel comparison methodologies such as those based on the theory of ``analysis by synthesis'' will be introduced. We will last discuss the real-world multimedia applications of IQA, and give a  list of open challenging problems, in the hope of encouraging more and more talented researchers and engineers devoting to this exciting and rewarding research field.
\end{abstract}

\maketitle

\section{Introduction}
Image quality assessment (IQA), a long-standing task in the field of image and multimedia processing, has evolved rapidly in the past two decades \cite{zhou06modern}, and has also gained increasing attention from both academic and industry for its broad applications. In this extended abstract, we plan to divide and introduce IQA in the following four parts: 
\begin{itemize}
    \item Subjective IQA, the most straightforward and reliable way of assessing perceptual quality by humans;
    \item Objective IQA, constructing computational models to automate the quality assessment process;
    \item IQA model comparison, quantifying the (relative) quality prediction performance of the competing models;
    \item IQA model applications, considering the particularities of different forms of multimedia data.
\end{itemize}

\section{Subjective IQA}
The goal of subjective IQA is to collect \textit{reliable} mean opinion scores (MOSs) from human subjects on the perceived quality of test images.
Several subjective methodologies have been standardized in the ITU-R and ITU-T recommendations \cite{bt2002methodology}, which can be broadly categorized into single-stimulus, double-stimulus, and multiple-stimulus methods. Take the single-stimulus absolute category rating (ACR) as an example. Each test image is rated individually using the labels ``bad'', ``poor'', ``fair'', ``good'', and ``excellent'', which  are translated to the values $1$, $2$, $3$, $4$, and $5$ when calculating the MOS. 
Along with the introduction of subjective experimental procedures, many important (but subtle) designs are also discussed.
\begin{itemize}
    \item Which subset of test images are ``ideal'' to choose from the web-scale unlabeled database for human annotation?
    \item How much instruction should be provided to subjects for more consistent and less biased MOS collection?
   \item What are the general guidelines to set up the experimental environment, especially the viewing conditions?
\end{itemize}
The immediate results of subjective experiments are human-labeled image quality databases, which monitor the progress of objective IQA.  For example, the LIVE dataset~\cite{sheikh2006statistical} marks the switch from distortion-specific to general-purpose IQA. The CSIQ dataset \cite{larson2010most} enables cross-dataset comparison. The TID2013 dataset~\cite{ponomarenko2013color} and its successor KADID-10K~\cite{lin2019kadid} expose the difficulty of IQA methods in generalizing to different distortion types. The Waterloo Exploration Database~\cite{ma17waterloo} tests model robustness to diverse content variations of natural scenes. The LIVE Challenge Database~\cite{ghadiyaram2015massive} probes the synthetic-to-real generalization,
which is further evaluated by the KonIQ-10K \cite{hosu2020koniq} and SPAQ~\cite{fang20perceptual} datasets. At the end of this part, we will give a brief overview of these datasets, and share our thoughts on creating better IQA databases in terms of mining hard and diverse images and collecting reliable MOSs.

\section{Objective IQA}

Objective IQA aims to develop computational algorithms that are capable of  providing consistent quality predictions with human data. These models can be mainly classified into two categories: full-reference (FR) and no-reference (NR, or blind) models. FR-IQA methods assume full access to a pristine undistorted image (also referred to as the reference image) for quality assessment of a ``distorted'' image. NR-IQA models, on the other hand, do not require any reference information. We will first discuss full-reference models, and start with THE default quality metric - mean squared error (MSE) that has dominated the field of signal processing for more than $50$ years \cite{wang2009mean}. We will revisit the limitations of MSE by hand-crafting its counterexamples intuitively. This motivates the development of the structural similarity (SSIM) index \cite{wang04image}, a award-winning and widely adopted perceptual quality model. Since its inception in 2004, the design philosophy underlying SSIM continues to impact the IQA field up to today. Among a myriad of existing IQA models, we will sample a few that we believe advance the field from at least one of the following aspects:
\begin{itemize}
\item More accurate IQA in terms of explaining human data in existing databases;
\item Color IQA that gives a better account for color perception of the human visual system;
\item Misalignment-aware IQA that does not require the reference and distorted images to be precisely aligned;
\item Texture-aware IQA that provides an efficient characterization of texture similarity;
\item IQA based on other design principles, e.g., information theoretic and data-driven approaches.
\end{itemize}
We will conclude the discussion of FR-IQA models by pointing out an embarrassing and common design flaw: many IQA models fail to satisfy the identity of indiscernibles\footnote{Give an FR-IQA model $D(\cdot)$, where a lower score indicating better predicted quality with a minimum of zero, and two images $x,y$, the identity of indiscernibles refers to $D(x,y) = 0 \Leftrightarrow x = y$.}, which has a strong implication that they are not suitable for perceptual optimization.

We then switch our attention to NR-IQA, which is more practical and challenging due to the lack of  reference information. We will first describe a widely accepted design principle based on natural scene statistics (NSS) \cite{wang2011reduced}. The underlying assumption is that a
measure of the destruction of statistical regularities of natural images provides a reasonable approximation to perceived visual quality. Both hand-crafted and learned NSS in spatial and frequency domain will be described. In particular, we would like to put more emphasis on one NR model, namely, the naturalness image quality evaluator (NIQE) \cite{mittal13making}, which has began to show its potentials in benchmarking image processing algorithms in real settings.


Limited by the expressiveness of hand-crafted features, NSS-based  approaches have been surpassed by data-driven NR-IQA models based on convolutional neural networks in recent years.
Patchwise training, transfer learning, and quality-aware pre-training are means of compensating for
the lack of human data. Apart from summarizing the specialized architectural designs, we plan to draw the audience's attention to the latest learning paradigms for NR-IQA, including
\begin{itemize}
    \item Unified learning for NR-IQA from multiple IQA databases simultaneously without additional subjective testing for perceptual scale realignment \cite{zhang21uncer};
    \item Active learning for NR-IQA by failure identification and model rectification \cite{wang2021troubleshooting};
    \item Continual learning for NR-IQA, where the model evolves with new data while being resistant to catastrophic forgetting of old data \cite{zhang2021continual}.
\end{itemize}

\section{IQA Model Comparison}
Conventional IQA model comparison generally follows a three-step approach. First, pre-select a number of images to form the test set. Second, collect the MOS for each image in the test set to represent its true perceptual quality. Third, rank the competing models according to their goodness
of fit (e.g., Spearman rank-order correlation coefficient) on the test set. The one with the best result is declared the winner. We will discuss the limitations of this conventional method in terms of the representativeness of test samples  and the risk of overfitting. We will then introduce a series of alternative IQA model comparison methods, including
\begin{itemize}
    \item Maximum differentiation (MAD) competition \cite{wang2008maximum}, automatically synthesizing images that are likely to falsify the IQA model in question;
    \item Group MAD (gMAD) competition \cite{ma20group}, a discrete instantiation of MAD that is more efficient and controllable; 
    \item Eigen-distortion analysis \cite{berardino2017eigen}, a method for comparing image representations in terms of their ability to explain perceptual sensitivity in humans;
    \item Comparison of IQA models for perceptual optimization of image processing systems \cite{ding21comparison}.
\end{itemize}
All the above-mentioned methods are based on the idea of ``analysis by synthesis'', which is rooted in the pattern theory by Ulf Grenander.
\section{IQA Model Applications}
It is highly nontrivial to apply IQA techniques in the field of multimedia due to substantially different data formats and  particularities \cite{wang2016objective}. Subject to the time constraint, we plan to present a few demonstrating examples, including
\begin{itemize}
    \item High-dynamic-range imaging \cite{laparra2016perceptual}, where the input and output images have different bit depths;
    \item Image fusion \cite{liu2011objective}, where input and output have different numbers of images;
    \item Color-to-gray conversion \cite{ma2015objective} and colorization, where input and output images have different color channels;
    \item Image retargeting \cite{fang2014objective}, where input and output images have different spatial resolutions;  
    \item Stereoscopic images \cite{wang15quality}, where binocular vision should be modeled;
    \item Omnidirectional images \cite{sui21perceptual}, where viewing behaviors may be indispensable for quality assessment; 
    \item Screen content images \cite{fang17objective}, where non-natural image statistics should be extracted;
    \item Natural videos (in the streaming setting), where the time dimension is added, leading to complex spatiotemporal distortions.
\end{itemize}
We will definitely point to the audience useful resources for IQA applications that have respectfully not covered. We will also cover some general and intuitive applications of IQA such as automatic hyperparameter adjustment and optimization of image processing algorithms. 

As a final remark, through this tutorial, we sincerely hope more and more talented researchers and engineers are willing to join us, contributing to this exciting and rewarding field.

\bibliographystyle{ACM-Reference-Format}
\input{reference.bbl}

\appendix









\end{document}